\documentclass{elsarticle}

\usepackage{booktabs} 
\usepackage{multicol}
\usepackage{subfig}
\usepackage{booktabs} 
\usepackage{amsmath}

\begin{document}

\begin{frontmatter}

\title{Single Image Dehazing Based on Generic Regularity}

\author{Kushal Borkar}
\ead{kushal.b15@iiits.in}
\author{Snehasis Mukherjee}
\address{IIIT SriCity}


\ead{snehasis.mukherjee@iiits.in}


\begin{abstract}
This paper proposes a novel technique for single image dehazing. Most of the state-of-the-art methods for single image dehazing relies either on Dark Channel Prior (DCP) or on Color line. The proposed method combines the two different approaches. We initially compute the dark channel prior and then apply a Nearest-Neighbor (NN) based regularization technique to obtain a smooth transmission map of the hazy image. We consider the effect of airlight on the image by using the color line model to assess the commitment of airlight in each patch of the image and interpolate at the local neighborhood where the estimate is unreliable. The NN based regularization of the DCP can remove the haze, whereas, the color line based interpolation of airlight effect makes the proposed system robust against the variation of haze within an image due to multiple light sources. The proposed method is tested on benchmark datasets and shows promising results compared to the state-of-the-art, including the deep learning based methods, which require a huge computational setup. Moreover, the proposed method outperforms the recent deep learning based methods when applied on images with sky regions.
\end{abstract}


\end{frontmatter}


\section{Introduction}
\label{sec:introduction}
Images captured by the camera, are often affected by haze due to several atmospheric conditions such as fog, smoke, multiple light sources, the scattering of light, etc. Presence of haze in an image obscure the clarity and hence needs to be restored. The flux of light per unit area received by the camera from the scene is attenuated along the line of sight. This redirection of light lessens the immediate scene transmission and replaces with a layer of scattered light known as airlight. Such scattering of light due to rigid particles floating in the air reduces the visibility of the scene.
\begin{figure}[t]
  \begin{multicols}{2}
	\subfloat[]{
	  \includegraphics[width=0.2\textwidth]{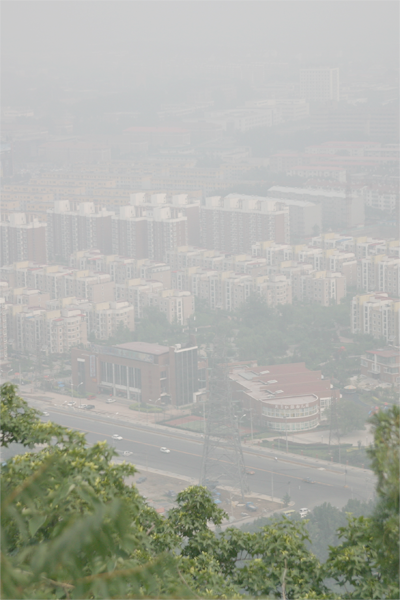} 
	  }
	\subfloat[]{
      \includegraphics[width=0.2\textwidth]{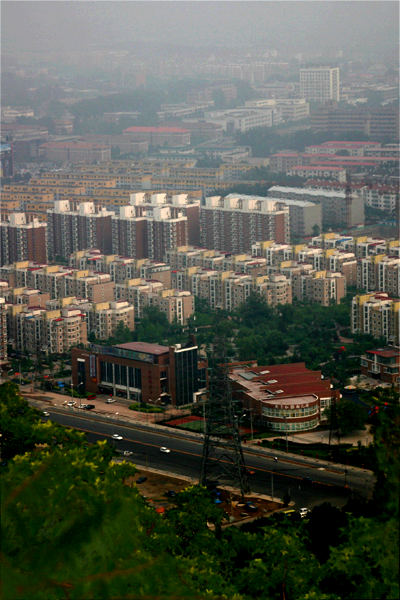} 
      }
  \end{multicols}
  \vspace{0pt}
 \caption{Sample result of the proposed approach. (a) Input hazy image. (b) Dehazed image using our approach.}
\end{figure}

Image methods endeavor to recover original scene radiance by expelling the impact of haze from the picture(as shown as an example in Figure 1). Since the effect of scattering of light at any pixel depends on the depth of the pixel, the degradation of the image due to haze is spatial-variant. However, recovering the effect of scene radiance from a pixel on an object is perplexing as the measure of fog and haze relies upon the separation between the object and the camera. In this way, global enhancement strategies do not function admirably for a scene image, because of the presence of objects at different depth and the background. The conventional methods exploited multiple images to overcome the problem. For example, the method of Narasimhan et.al. \cite{c1} require multiple pictures of a similar scene taken under various climate conditions. The method of Shwartz et.al. \cite{c2} requires images with the various level of polarization.
\begin{figure*}[!htbp]
  \begin{multicols}{4}
  \subfloat[]{
  \includegraphics[width=1\linewidth]{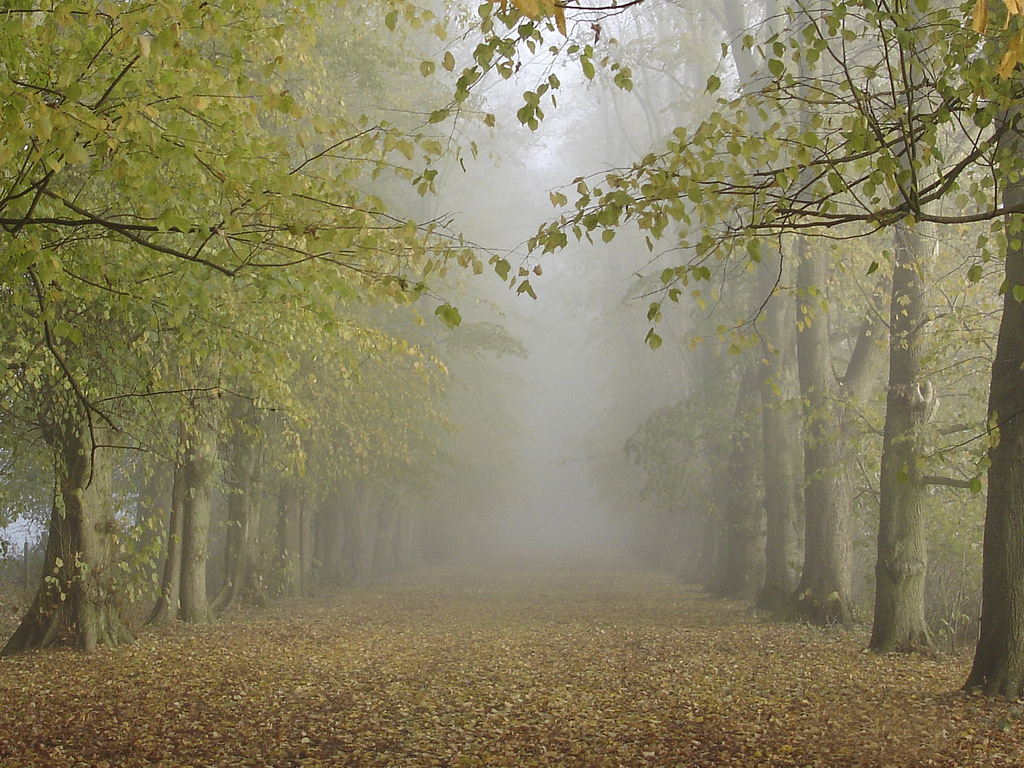}}
  \vspace{0.00mm}
  \subfloat[ ]{
  \includegraphics[width=1\linewidth]{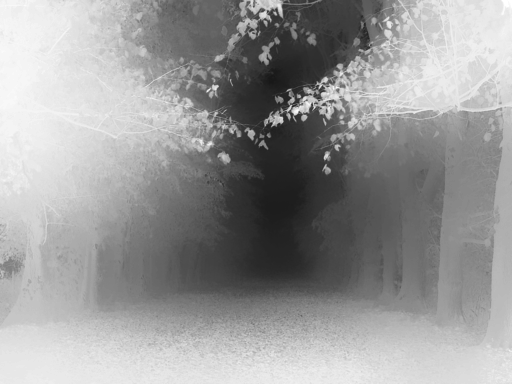}}
  \vspace{0.00mm}
  \subfloat[ ]{
  \includegraphics[width=1\linewidth]{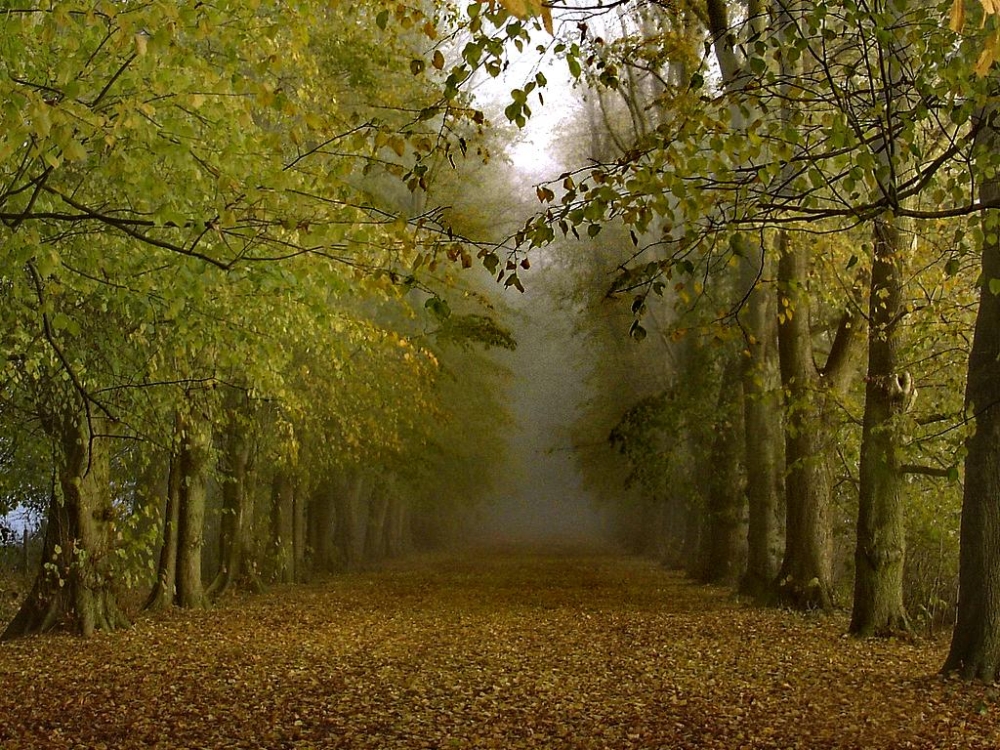}}
  \vspace{0.00mm}
  \subfloat[ ]{
  \includegraphics[width=1\linewidth]{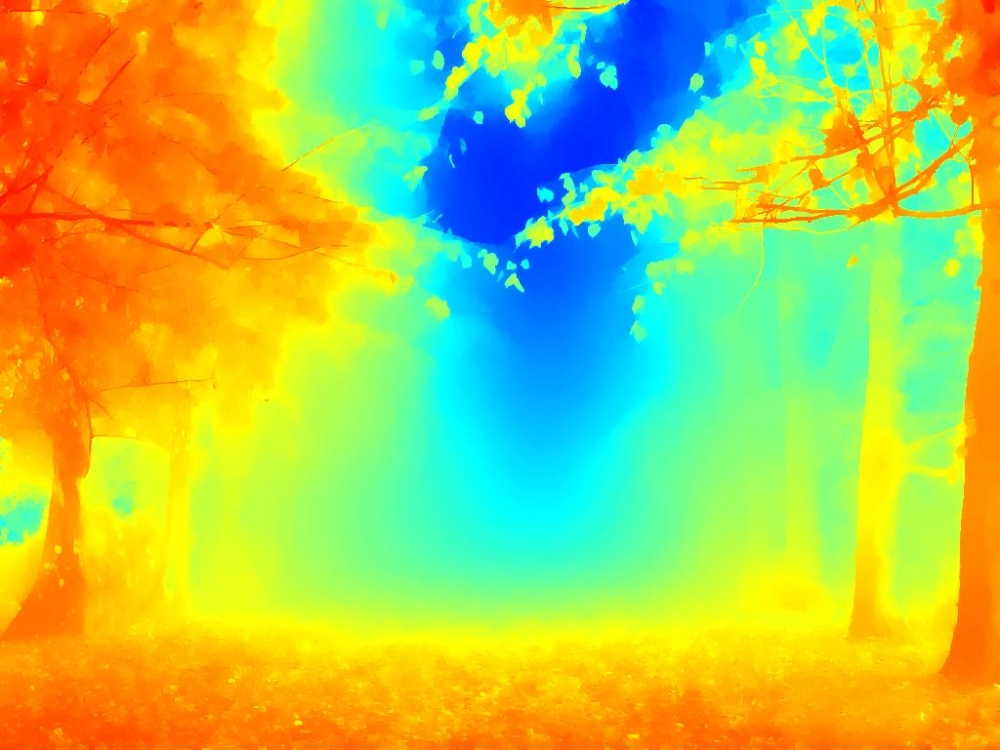}}
  \end{multicols}
\end{figure*}
\begin{figure*}
\vspace{0pt}
\caption{ Result of dehazing an image. (a) input haze image. (b) refined transmission map after NN regularization (c) final haze-free image (d) depth map of the output Haze-free Image.}
\end{figure*}

However, images taken in various climatic conditions may be unavailable always. Recently researches are going on for single image dehazing, where reference images for the given hazy images are unrequired.

After the introduction of the concept of Dark Channel Prior(DCP), people started dehazing an image without the help of a reference image \cite{c3}. Tan et.al. proposed a DCP based strategy for the single image by maximizing the local image contrast adopting a Markov Random Field (MRF) framework \cite{c3}. Fattal \cite{c4} used DCP to estimate the medium transmittance accepting the surface shading, furthermore, medium transmission capacities are locally measurably uncorrelated. He et al. \cite{c5} proposed the concept of DCP for estimating transmission map of the image. The DCP based approaches produce good result, but cannot handle locales where shading segment does not shift fundamentally contrasted with noise. Nair et al. \cite{nair} proposed a DCP based dehazing technique using a center surround filter to reduce the execution time. Another line of thought was introduced by Fattal in \cite{c6}, where the color line was used for dehazing. The color line based methods approximate transmission map based on the shift of the direction of atmospheric light from the origin. However, both the approaches fail to retain the original color of the objects in the hazy image.

Currently, some deep learning based methods have been proposed, producing much better results compared to the handcrafted features like DCP and color-line \cite{c25,c26,c27}. However, in case of images where sky background is present, the deep learning based methods fail to dehaze properly, as the learned features are not capable of fixing the haze at high depth areas of the image. In this paper, the contribution over the state-of-the-art is three-folds. Unlike the state-of-the-art methods, the proposed method for a single image assesses the benefit of the generic regularity of the image in which, pixels with moderate intensities of small image patches exhibit concordant distributions in RGB space. Our second contribution is to substitute the commonly used soft matting technique \cite{c23} in assessing the transmission map for haze removal, by first discovering some settled points in a maximum filter and then applying Nearest-Neighbor (NN) regularization. One example of using the transmission map in haze removal by the proposed method is shown in Figure 2. We utilize pixels of small image patches as a rule accompanying a one-dimensional dispersion in RGB space, to approximate the intensity at each patch and interject at places wherein the estimate is not generally unreliable. We do not consider the atmospheric light-weight. Finally, unlike the state-of-the-art methods, we do not consider and estimate the transmission of the medium. Alternatively, we approximate the additional airlight present in the image patch and eliminate that to clear the haze.

We interject and regularize the fractional evaluations of pixel intensity values into an entire transmission map, and outline an Adaptive Naive Bayes Classification approach to deal with pixels with a significant depth (such as sky region). Contrary to the conventional field models which comprise of normal coupling between adjacent pixels, we settle the transmission in isolated locales by fluctuating the number of pixels. We take the likelihood of more number of pixels to get the suitable pixel intensity value.

The rest of the paper is organised as follows. Section 2 describes the problem and a survey on the existing literature. Section 3 displays the mathematical and numerical foundation for the proposed strategy for dehazing. Section 4 weighs up the proposed algorithm. Experimental results and concluding comments are given in Sections 5 and 6 respectively.

\section{Related Work}
\label{sec:related_work}
Dehazing is a task of image remaking; the corruption of a hazy image is a direct result of the suspended particles in the turbid air. Single image dehazing is a challenging task which draws significant attention from researchers of the fields of Computer Graphics and Image Processing. The physical model often used to characterize the haze formation that causes degradation of an image, is known as Koschmieder's atmospheric scattering model \cite{c7}:
\begin{equation}
\label{e1}
\ I(x) = t(x)J(x) + (1 - t(x))A,
\end{equation}
where $I(x)$ is the observed intensity value of pixel $x$ in image $I$; $J(x)$ is the scene radiance of a haze-free image at $x$ and $t(x)$ is the medium transmission describing the portion of the light which reaches the camera without scattering. $A$ is the airlight, a global vector quantity describing the ambient light. The goal of haze removal task is to recover $J$, $A$ and $t$ from $I$.

In (\ref{e1}), the transmittance $t(x)$ does not change much for a given wavelength in a hazy image. The term $J(x)t(x)$ of (\ref{e1}) is termed as \textit{direct attenuation} \cite{c3}, which provides the information about the quantity of radiance received by the observer. The second part, called \textit{airlight}, provides an approximation of the measure of the atmospheric light included due to the dissipating course of the viewer.

Several attempts have been made for fog and haze expulsion in outdoor scene pictures \cite{survey}. Notwithstanding, the considerable significant advance in the research on haze removal methods was the idea of DCP. It is observed that, in the more considerable part of the local neighborhood regions which do not contain the sky, a few pixels in many instances get extremely low-intensity value at the minimum one color (RGB) channel (called "dark pixels"). In case of hazy image, the intensity of these dark pixels in the corresponding channel is contributed by the airlight. Along these lines, these dark pixels can explicitly give a precise estimation of the fog's transmission. In any case, the progression to reclassify the transmission approximation utilizing soft matting technique often leads to transmission underestimate.

When the density of haze varies smoothly in space, the effect of the homogeneous atmosphere transmission $t$ can be shown as:
\begin{equation}
\label{e2}
\ t(x) = e^{\beta d(x)},
\end{equation}
where $d(x)$ is the depth at pixel $x$ and $\beta$ is the scattering coefficient (in three dimensional space) of the atmosphere. So, (\ref{e2}) specifies the amount of radiance received by the observer, which is attenuated exponentially with the depth.

Several algorithms for image dehazing follow the image formation model (\ref{e1}) to dehaze images by recouping $J$. However, to regain haze-free image, both (\ref{e1}) and (\ref{e2}) are followed. First $J$ is obtained by subtracting a constant value corresponding to the dark pixels and then estimate the value of transmission $t$ precisely by assuming $A$ is known. Haze lessens the contrast in the picture, and different techniques depend on this perception for rebuilding. He \textit{et al.} \cite{c10} amplifies the contrast in each patch, while keeping up a global coherent image. This method enhanced the contrast for image dehazing and as a result, distant objects to the viewer seemed to be smooth and over saturated, causing the dehazed image look artificial. In \cite{c11}, Park \textit{et al.} evaluated the amount of haze in the image, from the contrast between the RGB channels, which diminishes as haze increments. This postulation of estimation of amount of haze is erroneous in the greay area. In \cite{c15}, Zhu \textit{et al.} assessed the haze in view of the perception that hazy regions are portrayed by high brightness and low saturation.

Some efforts have been made by utilizing a prior on the depth of the picture, for haze expulsion. A smoothness prior is utilized in \cite{c14}, expecting the image to be smooth except the pixels with depth discontinuities. Nishino et al. \cite{c12} expected the albedo and depth as factually free and together approximated them using priors on both. The albedo-prior accepted the distribution of gradients in images of characteristic scenes displaying a heavy-tail dispersion, and it is estimated as a summed up ordinary appropriation and further approximated as Gaussian distribution. The depth-prior is scene-dependent and is picked physically, either as piece-wise consistent for urban scenes or definitely changing for non-urban scenes.

The DCP based approaches \cite{c10} accept that within a small image patche there will be no less than one pixel with a dark color channel and utilize this negligible incentive as an approximate of the haze. This prior works extremely well, with the exception of in bright regions of the scene where intensity values of the considerable number of channels are likewise high. Hence, DCP cannot manage the sky area of the images in light of the fact that the dull channel pixels are potentially inaccessible in those splendid picture areas. Also, DCP based techniques are tedious because of the soft matting procedure involved in the method.

Recently, several techniques have been proposed to address the shortcoming of DCP \cite{c10}. In \cite{c13}, color ellipsoids are fitted in the RGB color space. These ellipsoids are utilized to arrange a unified approach to deal with the constraints of the DCP based techniques, and another strategy was proposed to assess the transmission in every ellipsoid. In \cite{c5,c16}, color lines were fit in RGB space, searching for little fixes with a steady transmission. The traditional color line based methods depend on the assumption that pixels in a haze-free image constitute color lines in RGB space \cite{c7}. These lines go through the inception and come from color varieties inside items. As appeared in \cite{c5}, the color lines in foggy scenes do not go through the starting point any longer, because of the added substance fog part.

Some significant attempts have been made using some priors other than the DCP and color line \cite{c17,c24,c28,c29}. In \cite{c28}, globally guided regularization technique is applied for dehazing. In \cite{c29}, contrast of the image is enhanced using boundary conditions. Tang et al. proposed a learning based technique for image dehazing \cite{c24}. However, the DCP and colorline based methods continue to dominate the other methods due to better performances, except images with high depth area (such as sky area). Riaz \textit{et al.} proposed a DCP based approach \cite{riaz} where the inefficacy of DCP at sky area is handled by limiting the contrast at the sky region. However, in most of the cases limiting the contrast causes loss of minute color information of the image.

The proposed method follows the concepts of DCP \cite{c10} and color line \cite{c5}. In addition to applying the above two priors, the proposed method applies a Nearest Neighbor (NN) regularization technique which helps reduce the execution time. The domain-adaptive nature of the patches helps in maintaining the original color and contrast of the image. Finally, the adaptive nature of the size of the mask help in maintaining the original color in high depth areas of the image (area which is far from the camera, such as, sky area).

Recently, deep learning based techniques are being successfully applied in all areas of Computer vision. A few efforts have been made to apply deep learning techniques for haze removal \cite{c26,c27,c25}. Unlike the traditional image dehazing techniques, AOD-Net in \cite{c27} proposes an end-to-end system for image dehazing, where a simple CNN is introduced to directly estimate the haze-free image, without obtaining the transmission map of the image. Dehazenet is proposed in \cite{c26}, where a transformation map of the input image is obtained from a CNN, by a trainable model. The CNN based methods usually give better results for image dehazing, compared to the traditional prior based models. However, for scene images where objects are far from the camera, the CNN based methods do not work well. Also, the deep network-based methods need a huge computational set up, which may not be affordable in many cases, e.g., hand-held devices. Next we discuss the mathematical background of the proposed method.

\section{Background of the Proposed Approach}
\label{sec:background_approach}
We assess the transmission by considering $A$ not constant for the scene. We observe that (\ref{e1}) expect the pixel intensity $I(x)$ to be radiometrically-straight. Hence, additionally to adjust methods that rely upon this course of action illustrate, our system requires the reversal of the procurement nonlinearities. In this section we explain the mathematical background of the proposed approach and the significance of the approach in the image dehazing task.

\subsection{Dark Channel Prior (DCP)}
\label{subsec:dcp}
The DCP is based on the perception on haze-free images: the most of the cases, the patches taken from a haze-free image must have no less than one color channel having a low intensity incentive at a few pixels \cite{c10}. According to this perception, for any image $J$, we characterize:
\begin{equation}
\label{e3}
J^{dark}(x) = \min_{c \in \{r,g,b \}}(\min_{y \in \Omega (x) } J^{c}(y)),
\end{equation}
where $\Omega (x)$ is the neighborhood patch centered at $x$, $J^c$ is the color channel of the picture $J$ and $J^{dark}$ the dark channel of $J$. The intensity of $J^{dark}$ is low and tends to zero if $J$ is a haze-free image. The low intensities in the haze-free locale are due to the following three elements:
\begin{itemize}
\item Shadows of objects: For example, the shadows of leaves, trees and rocks in landscape images or the shadows of buildings in cityscape images;
\item Dark objects or surfaces: For example, dark tree trunk and stone;
\item Colorful objects or surfaces: Color of any object (for example, green grass/tree/plant, blue water surface or red flower/leaf/wall) mostly tends to be very close to one of the color planes such as, Red, Green, Blue, Cyan, Magenta, Yellow, Black and White.
\end{itemize}
However, the DCP based techniques work well for images with high contrast and low depth region. For images consisting of regions with high depth value, DCP based methods fail to maintain the original color of the region. For example, sky region is usually bright and hence, do not have dark intensity values. We can deal with the sky locales by utilizing the Haze atmospheric scattering model (\ref{e1}), which is applied in the proposed model and discussed in the next subsection.
\begin{figure}[t]
  \begin{multicols}{4}
	\subfloat[]{
	  \includegraphics[width=0.225\textwidth]{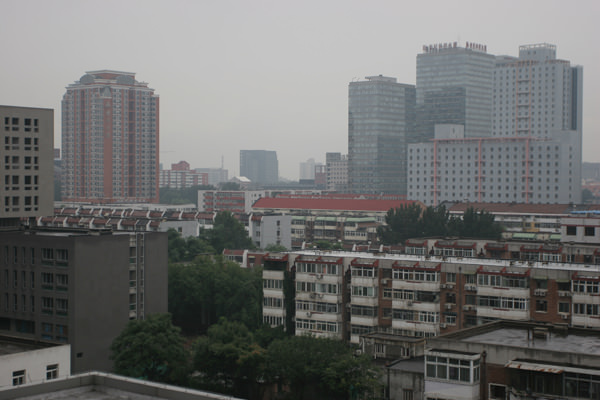} 
	  }
	\subfloat[]{
      \includegraphics[width=0.225\textwidth]{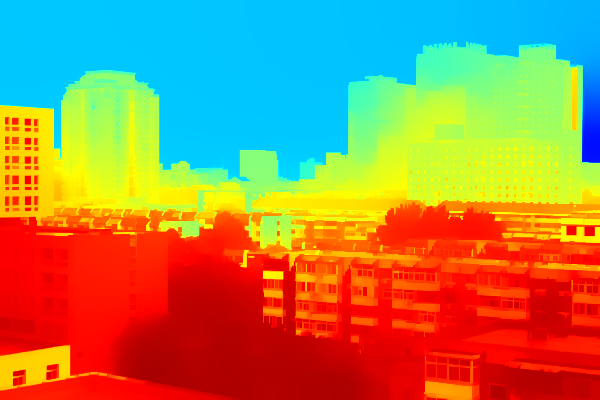} 
      }
  \subfloat[]{
	  \includegraphics[width=0.225\textwidth]{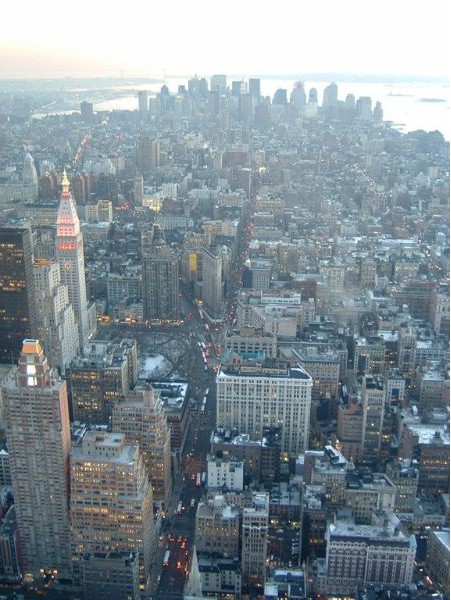} 
	  }
	\subfloat[]{
      \includegraphics[width=0.225\textwidth]{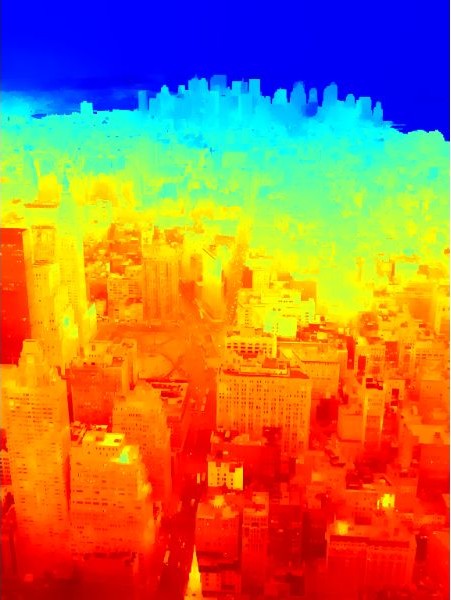} 
      }
  \end{multicols}
  \vspace{0pt}
 \caption{(a), (c) Input hazy images. (b), (d) Depth map of the Input Hazed Images}
\end{figure}

\subsection{Nearest-Neighbor (NN) Regularization}
\label{subsec:nn_regularization}
In Figure 3, we can observe that the transmission map is of an even consistency but is having unanticipated depth jumps in the hazy image. In the image-dehazing context, we assume that comparable pixels have an almost identical transmission value, according to \cite{c5,c18}.

In order to proficiently quantify the similarity between two image pixels, we have to think about the spatial variation and smoothness. Given an image pixel $x$, we characterize its element vector for similitude estimation in light of NN regularization as given below:
\begin{equation}
\label{e4}
f(x) = (R,G,B,\lambda X,\lambda Y)^T,
\end{equation}
where the intensities of $x$ are depicted as $R$, $G$ and $B$ in the RGB color space, respectively; $X$ and $Y$ are the spatial coordinates of $x$, and $\lambda$ is the balancing factor.

Despite the fact that we utilize indistinguishable features from \cite{c19}, however, is not the same as matting, so we adjust our technique to the important and broadly utilized conditions for matting utilizing (\ref{e4}).

\subsection{Color line model}
\label{subsub:color_line}
Colors in the small patches of a haze-free image generally lie on the line going through the starting point as shown by Omer \textit{et al.} \cite{c7}, if we represent colors as coordinates in the the RGB space. But the line is shifted from the origin by $A$ due to the additive airlight component in the case of hazy images. We assume that $A$ is constant. As we expect piecewise smooth depth $d$ which gives us a smooth scattering coefficient $\beta$ which verifies that $t(x)$ is piecewise smooth which is smooth at pixels that correspond to the same object. And $t(x)$ is also varying smoothly and slowing in the scene known from equation (\ref{e2}) except at depth discontinuities. This assumption of piecewise smooth geometry was used by Carr \textit{et al.} \cite{c32} and Nishino \textit{et al.} \cite{c12}. So, the equation (1) for a small patch can be rewritten as:
\begin{equation}
\label{e5}
I(x) = J(x)t + (1 - t)A, x \in \Omega,
\end{equation}
where $t$ is a fixed transmission value in the patch $\Omega$, and the airlight component is known to us. If we estimate the color line directly, then the measurement of color line direction will be erroneous if some dark pixels are present. In the proposed model, as the image is first subjected to DCP, will be less prone to error.

\subsection{Adaptive Naive Bayes Classification Model}
\label{subsub:naive_bayes}
Naive Bayes is a conditional likelihood model. Given an issue occurrence to be characterized, spoken to by a vector ${\displaystyle \mathbf {x} =(x_{1},\dots ,x_{n})}$ representing $n$ features (independent factors), it relegates to this case probabilities $p(C_{k} | x_{1},\dots ,x_{n})$ for each one of the $K$ conceivable results or classes ${\displaystyle C_{k}}$. The predicament with the above expression is that, if the quantity of highlights $n$ is substantial or if an element can go up against a considerable number of values, at that point applying such a model on likelihood tables is troublesome. We, therefore, reformulate the model to make it less demanding to deal with. Appropriating Bayes' theorem, the conditional likelihood can be calculated as:
\begin{equation}
\label{e6}
p(C_{k}  |  \mathbf{x}) = \frac{p(C_{k}) \cdot p(\mathbf{x} | C_{k})}{p(\mathbf{x})}.
\end{equation}
The denominator of (6) can be considered as a constant. The numerator is proportional to the conditional likelihood $p(C_{k} | x_{1},\dots ,x_{n})$, which can be revised as given below, utilizing the chain lead for numerous implementations of the meaning of conditional likelihood:
\begin{flalign}
p(C_{k},  x_{1},\dots ,x_{n}) &= p(x_{1},\dots ,x_{n}, C_{k}) \notag \\
&= p(x_{1}|x_{2},\dots ,x_{n}, C_{k}) \cdot  p(x_{2},\dots ,x_{n}, C_{k}) \notag \\
&= p(x_{1}|x_{2},\dots ,x_{n}, C_{k}) \cdot p(x_{2}|x_{3},\dots ,x_{n},\notag\\
& \quad C_{k}) \cdot p(x_{3},\dots ,x_{n}, C_{k})\notag\\
&= p(x_{1}|x_{2},\dots ,x_{n}, C_{k}) \cdot p(x_{2}|x_{3},\dots ,x_{n},\notag\\
& \quad C_{k}) \cdot p(x_{3}|x_{4},\dots ,x_{n}, C_{k})\notag\\
&= \dots \notag \\
&= p(x_{1}|x_{2},\dots ,x_{n}, C_{k}) \cdot p(x_{2}|x_{3},\dots ,x_{n},\notag\\
& \quad C_{k}) \cdots p(x_{n-1}|x_{n},C_{k}) \cdot p(x_{n}|C_{k}) \cdot p(C_{k})\notag \\
& \quad  
\end{flalign}
The ``naive" conditional independence presumptions are connected here for the proposed dehazing approach. Let us expect that each component ${\displaystyle x_{i}}$ is restrictively independent of each different feature ${\displaystyle x_{j}}$ for ${\displaystyle j\neq i}$, given the classification ${\displaystyle C}$. Which implies, $p(x_{i} | x_{i+1},\dots ,x_{n}, C_{k}) = p(x_{i} | C_{k})$. Along these lines, the proposed conditional likelihood model can be communicated as
\begin{flalign}
p(C_{k}  |  x_{1},\dots ,x_{n}) &\propto p(C_{k})p(x_{1}|C_{k})p(x_{2}|C_{k}) \dots \notag \\
&= p(C_{k}) \prod^{n}_{i=1} p(x_{i}|C_{k}).
\label{e8}
\end{flalign}
where $\propto$ denotes proportionality.

For evaluating the colorline heading, we develop a model that assigns the value in view of the component vector for similitude estimation in condition (4), where the element vectors are directly free. In attendance, we shift the quantity of pixels given to the Naive Bayes Classification Model. By changing the number of pixels we take the probability of more number of pixels by finding the probabilities utilizing the Naive Bayes' condition to compute the posterior probability for every pixel based on the inherent vector in (4). The pixel with the most astounding posterior probability obtain the result of the proposed expectation model.

\section{Proposed Approach}
\label{sec:our_approach}
In this section, we explain the steps to dehaze an image utilizing the nearby fix demonstrate in (\ref{e5}) and its related transmission estimation system in (\ref{e2}). We started with a brief review of the mathematical background in the previous section. Now we demonstrate the way the mathematical models can be used to check the input image and consider small patches of pixels as possible fixes that satisfy (\ref{e3}). As articulated in the previous section, pixels that relate to an almost planar surface lie on a color line in RGB space depicted by (\ref{e4}) and (\ref{e5}). In this way, in each fix we run a Naive Bayes Classification Model unequivocally built using (\ref{e8}) that scans for a line bolstered by a significant number of pixels. We at that point check whether the line found is reliable with our arrangement show by testing it against a rundown of conditions postured by the model. A line that concludes everyone of these tests effectively is then utilized for assessing the transmission as per (\ref{e3}) and (\ref{e4}). The subsequent values at that point are allocated to every one of the pixels that help in finding the color line. We evaluate the transmission in patches where we neglect to discover a line that meets every one of the conditions by utilizing the NN Regularization and reapplying the above strides on the relating patch. Next we discuss the process of estimating transmission map of an image using DCP.

\subsection{Transmission Estimation Using Dark Channel Prior}
\label{subsec:transmission_calc_dcp}
We believe that the transmission $t$ within a neighborhood patch $\Omega(x)$ is steady. Performing the minimum operation in the neighborhood patch of the hazy picture in (\ref{e4}), we have the following:
\begin{equation}
\label{e9}
\min_{y \in \Omega (x)}(I^c (y)) = \min_{y \in \Omega (x) }(J^c (y))t + (1 - t)A^c.
\end{equation}
The above equation can be derived further as:
\begin{equation}
\label{e10}
\min_{y \in \Omega (x) }(\frac{I^c (y)}{A^c}) = \min_{y \in \Omega (x) }(\frac{J^c (y)}{A^c})t + (1 - t).
\end{equation}
Considering the min operation again on the above condition, in order to get the minimum of the three colors in RGB channel, we have:
\begin{equation}
\label{e11}
\min_{c}(\min_{y \in \Omega (x)} (\frac{I^c (y)}{A^c}) = \min_{c}(\min_{y \in \Omega (x) }(\frac{J^c (y)}{A^c})t) + (1 - t).
\end{equation}
As indicated by (\ref{e3}), the dark channel $J^{dark}$ of the haze-free pixel intensity $J$ should tend to zero. Also, as $A^c$ is a positive constant, thus:
\begin{equation}
\label{e12}
\min_{c}\Big(\min_{y \in \Omega (x) }(\frac{J^c (y)}{A^c})\Big) = 0.
\end{equation}
From (\ref{e9}) and (\ref{e10}), we get:
\begin{equation}
\label{e13}
\ t = 1 - \min_{c}\Big(\min_{y \in \Omega (x) }(\frac{I^c (y)}{A^c})\Big).
\end{equation}
The above mathematical formulation gives us an estimation of the transmission. As stated earlier, the DCP is not a appropriate prior for the sky locales. Since the sky is at boundless infinite depth and practically has zero transmission, the (\ref{e9}) smoothly tackles both sky and non-sky regions in the proposed approach. We dehaze the image from the assessed approximation of transmission obtained from (\ref{e9}) and utilize the neighborhood local patch model.

Reasonably, even in clear environment, the atmosphere around us is not completely free of any molecule. Along these lines, the haze still exist when we have a look at inaccessible items. In addition, the presence of haze is a major prompt for a human to comprehend depth.

\subsection{Dehazing Algorithm}
\label{subsec:algorithm}
In this component, we illustrate the initiatives for dehazing an image using our method. First, we dehaze the image by removing the dark pixels from the image without applying a soft matting algorithm \cite{c23} to convalesce the transmission. We signify the transmission outline $t(x)$ and get the accompanying cost function:
\begin{equation}
\label{e14}
E(t) = t^TLt + \lambda (t - \tilde{t})^{T+1},
\end{equation}
where $L$ is the Matting Laplacian matrix introduced by Levin \cite{c8}, and $\lambda$ is a regularization parameter.

We analyse the transmission estimations of a few points with low depth values, and propose recuperating the transmission estimations of the rest of the points by finding their closest match from the arrangement of exact focuses in light of the built k-d tree with the 5-dimensional component vectors characterized in (\ref{e4}). Conforming transmission maps along with the input image is shown in Figure 3, from which we can perceive that the transmission outline is significantly smooth aside from unanticipated depth jumps. As specified earlier, the dehazing issue is extremely under-constrained. Hence, we have to surmise a few assumptions from (\ref{e8}), (\ref{e11}) and (\ref{e12}) on the normal transmission map. The refined transmission outline exercising the above strategy accomplishes to capture the sharp edge discontinuities and outline the shape of the items. With the transmission outline, we can remunerate the scene radiance as per (\ref{e1}). Yet, the direct attenuation may be near to zero when the transmission is near to zero. In this way, we limit the transmission $t(x)$ to a lower bound $t_0$, so that a specific measure of haze are retained. The scene radiance $J(x)$ is eventually recouped by:
\begin{equation}
\label{e15}
J(x) = \frac{I(x) - \hat{A}}{max(t(x),t_{0})} + \hat{A},
\end{equation}
where $\hat{A}$ denotes the estimated airlight of the image.

A traditional range of $t_{0}$ falls in the interval [0.09, 0.1], which is fixed experimentally. Since the scene radiance is generally unlike the atmospheric light, the image after disturbance by haze evacuation looks diminish. Hence, we increse the insight into $J(x)$ for illustration. We examine the information from the input image and consider limited windows of pixels as hopeful patches following (\ref{e4}). Pixels relating to an almost-planar mono-chromatic surface lie on a color line in RGB space as indicated by (\ref{e4}). We evaluate the color line by applying ``Naive Bayes Classifier" on the intensity values in RGB space and get two focuses $\vec{p_{1}}$ and $\vec{p_{2}}$ lying on the straight line and an arrangement of inliers. The condition of the line will be $\vec{P} = \vec{P_{0}} + \rho \vec{D} $, where $\rho$ represent the parameter of the line, $\vec{P_{0}} = \vec{p_{1}}$ and the heading proportion $\vec{D} = \frac{\vec{p_{2}} - \vec{p_{1}}}{\mid\mid ( \vec{p_{2}} - \vec{p_{1}} )\mid\mid}$. We examine the assessed line with conditions in particular, noteworthy number of inliers, positive incline of $\vec{D}$ and unimodality analogously in \cite{c5} and eliminate the questionable ones. Figure 4 demonstrates the effect of applying the Naive-Bayes Classifier and the proposed NN regularization technique on a sample hazy image. The second row of Figure 4 illustrates the effect of applying different patch sizes on the same image.
\begin{figure*}[!htbp]
  \begin{multicols}{3}
  	\subfloat[]{
  	\includegraphics[width=1.09\linewidth]{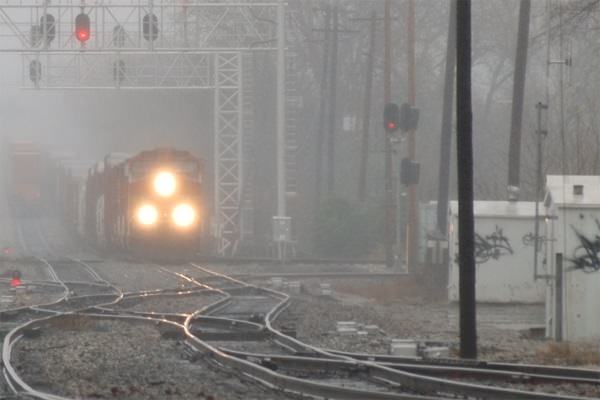}}
  	\subfloat[ ]{
  	\includegraphics[width=1.09\linewidth]{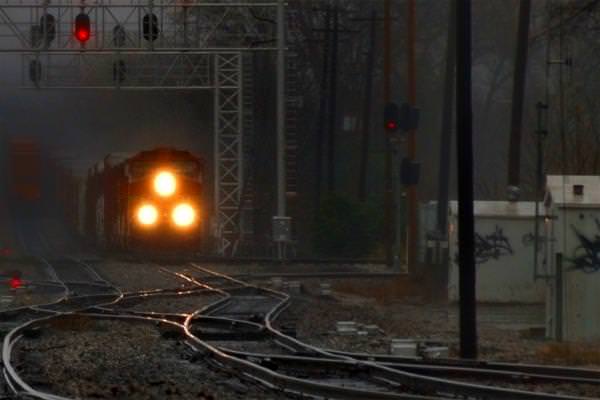}}
  	\subfloat[ ]{
  	\includegraphics[width=1.09\linewidth]{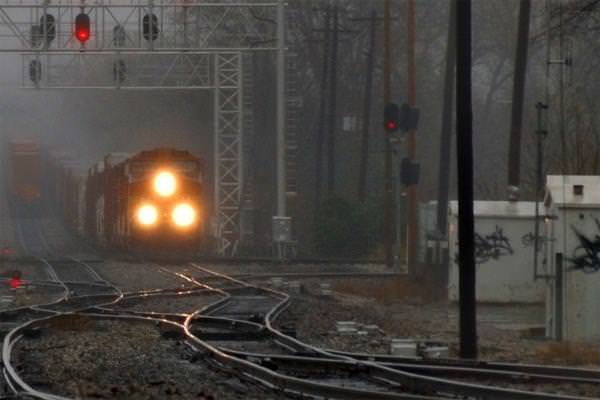}}
  \end{multicols}
  \begin{multicols}{3}
  	\subfloat[]{
  	\includegraphics[width=1.09\linewidth]{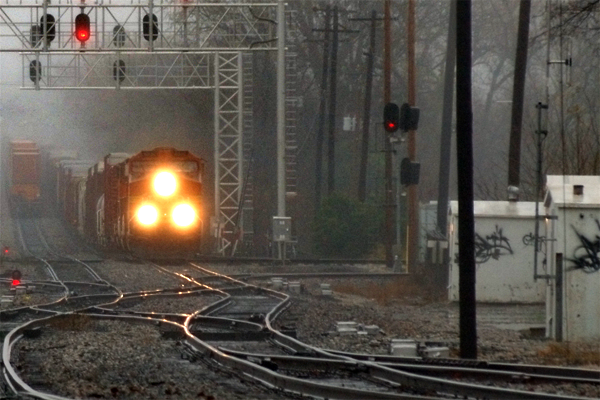}}
  	\subfloat[ ]{
  	\includegraphics[width=1.09\linewidth]{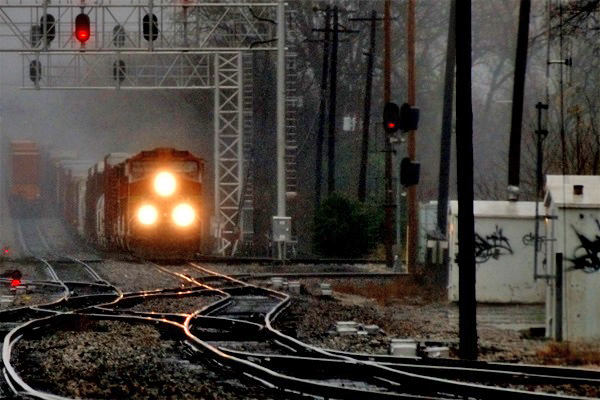}}
  	\subfloat[ ]{
  	\includegraphics[width=1.09\linewidth]{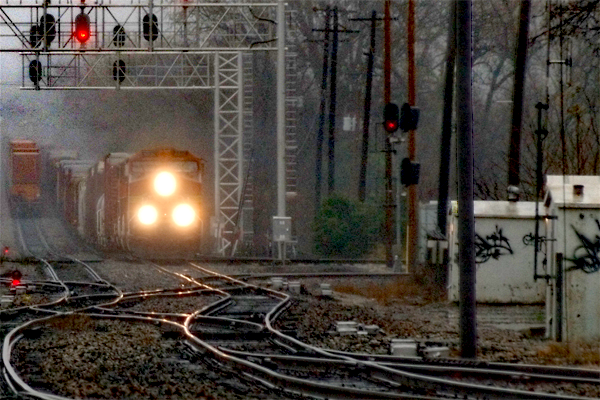}}
  \end{multicols}
\end{figure*}
\begin{figure*}
\vspace{0pt}
\caption{Result of the proposed method with varying patch size. (a) input haze image. (b) Output Image without Nearest Neighbour Regularization (c) Output image without applying Naive Bayes' Classification Model (d) pixels are given to Naive Bayes Classification Model with patch size 7 $\times$ 7 (e) 15 $\times$ 15 and (f) 30 $\times$ 30.}
\end{figure*}

\subsubsection{Computing $\hat{A}$}
\label{subsubsec:compute_A}
We can make our estimates for $\hat{A}$ using the Dark Channel \cite{c3} of the patch. Dark Channel of the specific patch $\Omega$ is determined as follows (following \cite{c34}:
\begin{equation}
\label{e16}
Dark(\Omega) = \min_{x \in \Omega (x) }\Big(\min_{c \in {R,G,B}} I_c(x)\Big),
\end{equation}
where $I_c$ is the $c$th color channel, for $c\in \{R,G,B\}$.

We have already defined the direction ratio $\vec{D}$ in the previous section, considering the standard conditions, significant line support, positive reflect of $\vec{D}$, unimodality and valid transmission according to (\ref{e13}) and (\ref{e16}) as discussed in \cite{c5}. We check whether the visible pixels efficiently found are consistent with our formation model by running it following (\ref{e4}), (\ref{e15}) and (\ref{e16}). With respect to $\vec{D}$, we conventionally consider the normal to the inclined plane from the precise origin as determined as $\vec{N} = \frac{\vec{p_{2}} \times \vec{p_{1}}}{\mid\mid ( \vec{p_{2}} \times \vec{p_{1}} )\mid\mid}$.

We compute $\hat{A}$ by minimizing the following error from the existing corresponding normals from the above defination:
\begin{equation}
\label{e17}
E(\hat{A}) = \frac{1}{\Omega} \cdot \sum_{i \in {\Omega}} (N_i \cdot \hat{A}),
\end{equation}
where the line parameters are denoted by $N_i \cdot \hat{A}$ of the patch pixels. The $\cdot$ typically denotes the dot-product in RGB space. This reliable measure consists of projecting the line parameters onto an error function, $E(\hat{A})$. Therefore, (\ref{e17}) vanishes over uniformly distributed pixels and becomes positive. Therefore, we have
\begin{equation}
\label{e18}
\frac{\partial E}{\partial \hat{A}} = \hat{A} \cdot \Big(\sum_{i} N_iN_i^T \Big) = 0.
\end{equation}

From the (\ref{e18}), we need the non-trivial solution as $\hat{A}$ is a non-null vector which can be acquired by registering a covariance matrix from the normals and afterwards obtaining the result as a covariance matrix similar to the smallest eigenvalue.

\subsubsection{Estimating the Magnitude of Airlight}
\label{subsubsec:magnitude_of_airlight}
We can retrieve the magnitude $a(x)$ of the airlight component, from the evaluated $\hat{A}$ and acquire the color line correlating with the patch by limiting the accompanying error:
\begin{equation}
\label{e19}
\ E_{line} (\rho, s) = \min_{\rho, s} \mid\mid {P_{0}} + \rho D - s \hat{A} \mid\mid ^2.
\end{equation}
Here, airlight component is denoted by $s$. The solution of  (\ref{e19}) is obtained following \cite{c6}. The registered airlight portion is then approved with the accompanying conditions: vast intersection and convergence angles, close intersection, substantial range and shading changeability. Out of them, vast intersection angle and close convergence are done likewise approaches as detailed in \cite{c6}.

We estimate $J$ from the refined transmission map and the output image obtained from the proposed method and a black image $E$. From the intermediate outcomes, it can be observed that the underlying $J$ has noticeable artefacts in the sky locale, which is progressively exterminating to amid the improvement. The outcome is given in Figure 5, exhibiting the gradual decline in the graph alongside the images as output.
\begin{figure*}[!htbp]
	\includegraphics[width=1.04\linewidth]{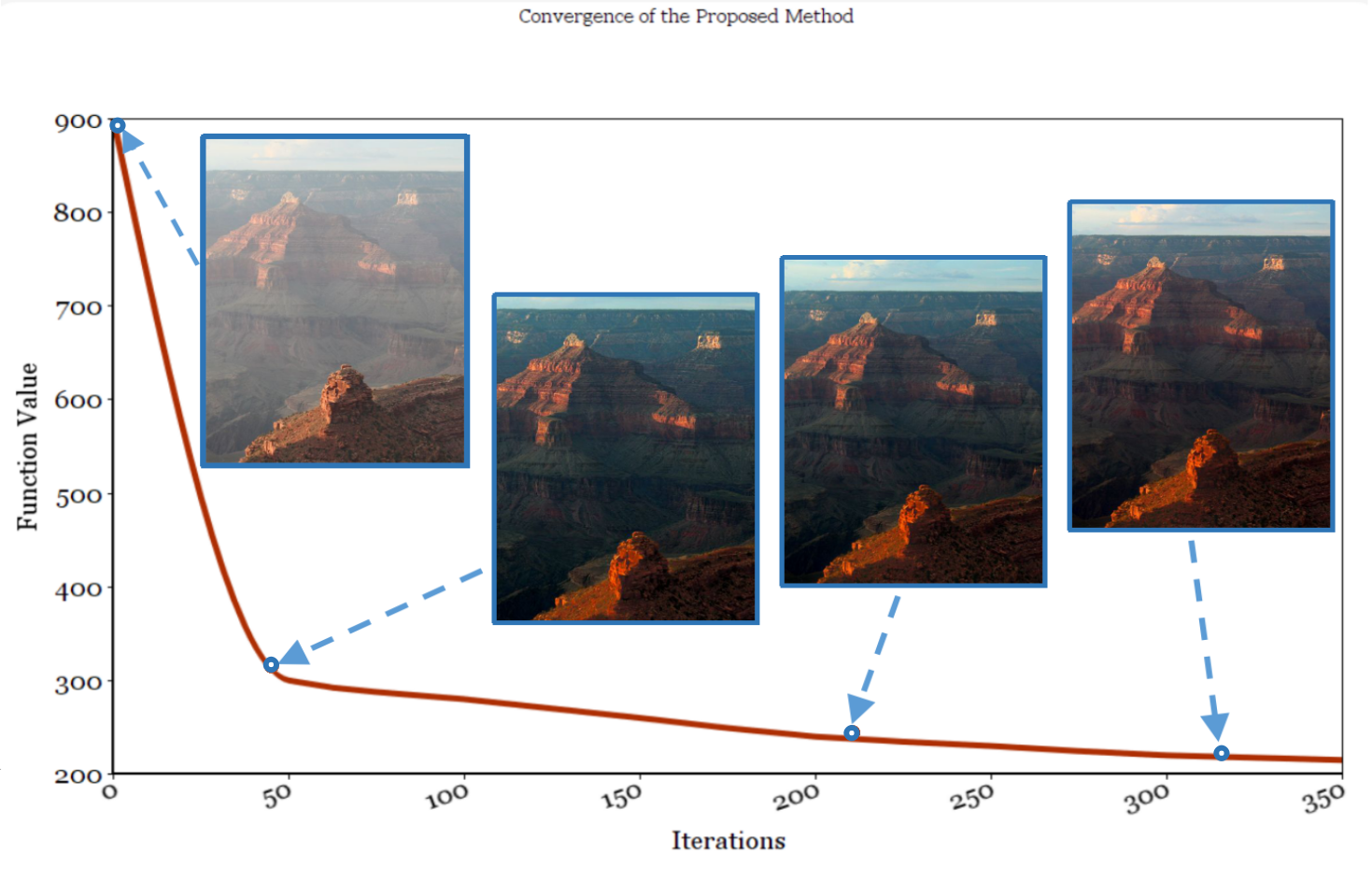}
	\caption{The convergence of the proposed approach. The objective function in Eq. (\ref{e19}) is monotonically declining. The transitional consequences of $J$ and $10 \times E$ at the given iteration.}
\end{figure*}

\subsubsection{Interpolation of Estimate}
\label{subsubsec:interpolation_of_airlight}
For assessing the airlight part $a(x)$ we dispose of a couple of patches and  introduce the estimation of airlight at each pixel. This is done by limiting the accompanying capacity:
\begin{equation}
\label{e20}
	\begin{aligned}
		\psi(a(x)) = \sum_{\Omega} \sum_{x \in {\Omega}} \frac{(a(\textbf{x}) - \widetilde{a} (\textbf{x})^2}{(\sigma_a (\Omega))^2} + \beta \cdot \sum_{x} \frac{a(\textbf{x})}{\mid\mid I(\textbf{x})\mid\mid}  \\
		+ \alpha \cdot \sum_{\Omega} \sum_{x \in {L(x)}} \frac{(a(\textbf{x}) - \vec{a} (\textbf{x})^2}{\mid\mid I(\textbf{x}) - I(\textbf{y}) \mid\mid^2}.
	\end{aligned}
\end{equation}
Here, $\vec{a}(x)$ is the assessed magnitude of the airlight component, $L(x)$ is the local neighborhood of $\textbf{x}$ and $a(\textbf{x})$ represents the adequate segment to be registered and $\sigma_a(\Omega)$ is the error difference of the estimate of $\vec{a}(x)$ inside the patch $\Omega$. The initial two terms constitute the capacity utilized as a part of \cite{c5}. To obtain a better result, we include the last term which guarantees that the part would be a little portion of $I(\textbf{x})$. The latter term of (9) is utilized to change the segment with the intensity of $I(\textbf{x})$. Presently, to limit the vitality work given by condition (20), we change over this to the following structure:
\begin{equation}
\label{e21}
\Psi(a) = (a - \widetilde{a})^T \cdot \Sigma(a - \widetilde{a}) + \alpha a^T La + \beta b^T a,
\end{equation}
where $a$ and $\widetilde{a} $ denote the vector forms of $a(\textbf{x})$ and $\widetilde{a}(\textbf{x})$ respectively, $\Sigma$ denote the covariance grid of the pixels where the approximation is made and $L$ is the Laplacian framework of the diagram built by considering each pixel as a vertex and interfacing the neighboring vertices. The influence of the edge between the pair of vertices $x$ and $y$ is $\frac{1}{\mid\mid I(\textbf{x}) - I(\textbf{y})\mid\mid^2}$. Here $\alpha$ and $\beta$ are scalars controlling the significance of each term.

\subsubsection{Haze Removal}
\label{subsubsec:haze_remove}
Airlight at every pixel is now obtained by figuring $a(\textbf{x})\hat{A}$. Consequently, the immediate transmission can be recouped from
\begin{equation}
\label{e22}
J(\textbf{x}) t(\textbf{x}) = I(\textbf{x}) - a(\textbf{x})\hat{A}.
\end{equation}
As we do not have $t(x)$ unequivocally, we elevate the contrast utilizing the airlight and attempt to recoup $J(\textbf{x})$. For instance, if the recouped image is $R_im (\textbf{x})$, at pixel $x$:
\begin{equation}
\label{e23}
R_im (\textbf{x}) = \frac{J(\textbf{x}) t(\textbf{x})}{1 - Y(a(\textbf{x})\hat{A})},
\end{equation}
where $Y$ is given by the following equation:
\begin{equation}
\label{e24}
Y(I(\textbf{x})) = 0.2989 I_R (\textbf{x}) + 0.5870 I_G (\textbf{x}) + 0.1140 I_B (\textbf{x}),
\end{equation}
where $Y(I(\textbf{x}))$ processes the luma at the pixel $x$. The idea is to upgrade the pixel $x$ relying upon how much intensity (brightness) is expelled out from it. In many cases, the image remains dull even after the above operation. So we utilize gamma revision to reestablish the natural shine of the image.

\section{RESULTS}
\label{sec: result}
We assess the performance of the proposed method on the Berkeley Segmentation Dataset (BSDS300), containing natural images \cite{c33}. This is a diverse dataset of clear open air characteristic images and consequently represents the moderate scenes that may have been affected by haze. We have evaluated airlight part in condition (17). We utilize similar parameters for every images: in condition (10) we set $\lambda$ = 0.1 and we scale $\frac{1}{\sigma ^2 (x)}$ in the interim [0,1] to maintain a strategic distance from numeric issues. For experimenting on the synthetic dataset, we utilize the dataset proposed in \cite{c5}. The dataset contains eleven dehazed pictures, manufactured distance maps and relating reenacted fog pictures. An indistinguishably appropriated zero-mean Gaussian commotion with three one of a kind clamour level: n = 0:01; 0:025; 0:05 was added to these images (with picture control scaled to [0; 1]).
\begin{table}
\label{tab1}
\captionof{table}{L1 errors (scaled to the interval [0,1]), after applying the proposed method over synthetic hazy images with different measure of noise and for different standard deviation; compared to the competing methods.}
\begin{tabular}{|c|c|c|c|c|c|}
\hline
\ & \textit{$\sigma$} & \cite{c5} & \cite{c10} & \cite{c19} & ours\\
\hline
\ & 0 & 0.097/ & 0.069/ & 0.058/ & \textbf{0.056}/ \\
\ & & 0.051 & \textbf{0.033} & 0.040 & \textbf{0.033}\\

\ & 0.01 & 0.100/ & 0.068/ & 0.061/ & \textbf{0.060}/ \\
\ & & 0.058 & \textbf{0.038} & 0.045 & 0.039\\
\ Road1 & 0.025 & 0.106/ & 0.084/ & \textbf{0.072}/ & \textbf{0.072}/\\
\ & & 0.074 & 0.065 & 0.064 & \textbf{0.063}\\
\ & 0.05 & 0.136/ & 0.120/ & \textbf{0.091}/ & 0.095/\\
\ & & 0.107 & 0.114 & \textbf{0.100} & 0.105\\

\hline

\ & 0 & 0.118/ & 0.077/ & 0.032/ & \textbf{0.030}/ \\
\ & & 0.063 & 0.035 & 0.026 & \textbf{0.020}\\

\ & 0.01 & 0.116/ & 0.056/ & 0.032/ & \textbf{0.029}/ \\
\ & & 0.067 & 0.038 & \textbf{0.032} & \textbf{0.032}\\
\ Lawn1 & 0.025 & 0.109/ & 0.056/ & \textbf{0.052}/ & \textbf{0.052}/\\
\ & & 0.077 & 0.065 & 0.056 & \textbf{0.052}\\
\ & 0.05 & 0.115/ & 0.114/ & 0.099/ & \textbf{0.095}/\\
\ & & 0.102 & 0.121 & 0.107 & \textbf{0.104}\\

\hline
\ & 0 & 0.074/ & \textbf{0.042}/ & 0.080/ & 0.050\\
\ & & 0.043 & \textbf{0.022} & 0.049 & \textbf{0.022}\\
\ & 0.01 & 0.067/ & 0.048/ & 0.088/ & \textbf{0.045}\\
\ & & 0.040 & \textbf{0.030} & 0.056 & 0.040\\
\ Mansion & 0.025 & \textbf{0.057}/ & 0.065/ & 0.104/ & 0.060\\
\ & & \textbf{0.044} & 0.051 & 0.072 & \textbf{0.044}\\
\ & 0.05 & 0.083/ & \textbf{0.081}/ & 0.116/ & 0.082\\
\ & & \textbf{0.075} & 0.080 & 0.095 & \textbf{0.075}\\

\hline
\ & 0.01 & 0.07 & 0.048 & 0.088 & \textbf{0.045} \\
\ & & 0.048 &\textbf{0.025} & 0.032 & \textbf{0.025} \\
\ & 0.01 & 0.067/ & 0.053/ & 0.049/ & \textbf{0.045} \\
\ & & 0.050 & 0.043 & 0.041 & \textbf{0.040}\\
\ Church & 0.025 & 0.058/ & 0.089/ & 0.047/ & \textbf{0.045}\\
\ & & 0.059 & 0.081 & 0.057 & \textbf{0.050}\\
\ & 0.05 & 0.087 & 0.121 & \textbf{0.043} & 0.047 \\
\ & & 0.121 & 0.136 & 0.092 & \textbf{0.85} \\
\hline
\end{tabular}
\end{table}

\subsection{Quantitative Analysis}
\label{sec:quantitative_analysis}
Table I condenses the L1 errors on non-sky pixels of the transmission maps and the fog free images of the synthetic dataset. The results of four sample images are given in the table. We utilize a progression of assessment criteria as distant as to separate between each coordinate of the haze-free image along with the result. Notwithstanding the extensively used mean square error (MSE) and the structural similarity (SSIM) \cite{c30} measures, we utilized some more assessment frameworks, for example, weighted peak signal-to-noise ratio (WSNR) and peak signal-to-noise ratio (PSNR) \cite{c31}. Our method is compared with seven state-of-the-art methods (including deep learning based methods) when applied on the  hazy images of the BSDS dataset, results of which is shown in Table II.
\begin{table}
\label{tab2}
\caption{The average results of MSE, SSIM, PSNR and WSNR on the hazy Images.}
\begin{tabular}{c| c c c c}
\hline \hline
\textit{Methods} & \textit{MSE} & \textit{SSIM} & \textit{PSNR} & \textit{WSNR} \\
\hline
\textit{Hazy Input Image} & 0.0481 & 0.9936 & 61.5835 & 8.5958 \\
\textit{He \cite{c10}} & 0.0172 & 0.9981 & 66.7392 & 13.8508 \\
\textit{Berman \cite{c20}} & 0.0201 & 0.9975 & 66.5450 & 13.7236 \\
\textit{Liao \cite{c29}} & 0.0120 & 0.9956 & 68.7843 & 12.6230 \\
\textit{Tang \cite{c24}} & 0.0070 & 0.9989 & 70.0099 & 17.1180 \\
\hline
\textbf{\textit{our result}} & 0.0067 & 0.9989 & 69.9867 & 18.0032 \\
\hline
\textit{MSCNN \cite{c25}} & 0.0075 & 0.9983 & 69.9767 & 17.7839 \\
\textit{DehazeNet \cite{c26}} & 0.0062 & 0.9993 & 70.9767 & 18.0996 \\
\textit{AODNet \cite{c27}} & 0.0064 & 0.9995 & 71.0007 & 18.0667 \\
\hline \hline
\end{tabular}
\end{table}

In Table II, the last three lines demonstrate the results of applying deep learning based techniques. As we can observe, the effects of applying the proposed method on hazy images is much better than the best in class handmade approaches. Additionally, the proposed method is comparable to the deep-learning based procedures as well.
\begin{table*}[t]
\begin{center}
\caption{The average results of MSE, SSIM, PSNR and WSNR on the Images with the Sky Region.}
\begin{tabular}{c|c c c c c c c c}
\hline \hline
 \textit{Metric} & \textit{Hazy} &  \cite{c10} & \cite{c24} & \cite{c25} & \cite{c26} & \cite{c27} & \textit{our result}\\
\hline
 \textit{MSE} &  0.0481 & 0.0172 & 0.0070 & 0.0075 & 0.0062 & 0.0064 & 0.0067 \\
\hline
 \textit{SSIM} & 0.9936 & 0.9981 & 0.9989 & 0.9883 & 0.9993 & 0.9995 & 0.9989 \\
\hline
\textit{PSNR} & 61.5835 & 66.7392 & 70.0099 & 69.9767 & 70.891 & 71.00 & 70.967 \\
\hline
\textit{WSNR} & 8.5958 & 13.8508 & 17.1180 & 17.7339 & 18.0886 & 18.0667 & 18.0032 \\
\hline \hline
\end{tabular}
\end{center}
\end{table*}

\begin{figure*}
\label{comp_graph}
\centering{
	\includegraphics[height=2in,width=2.3in]{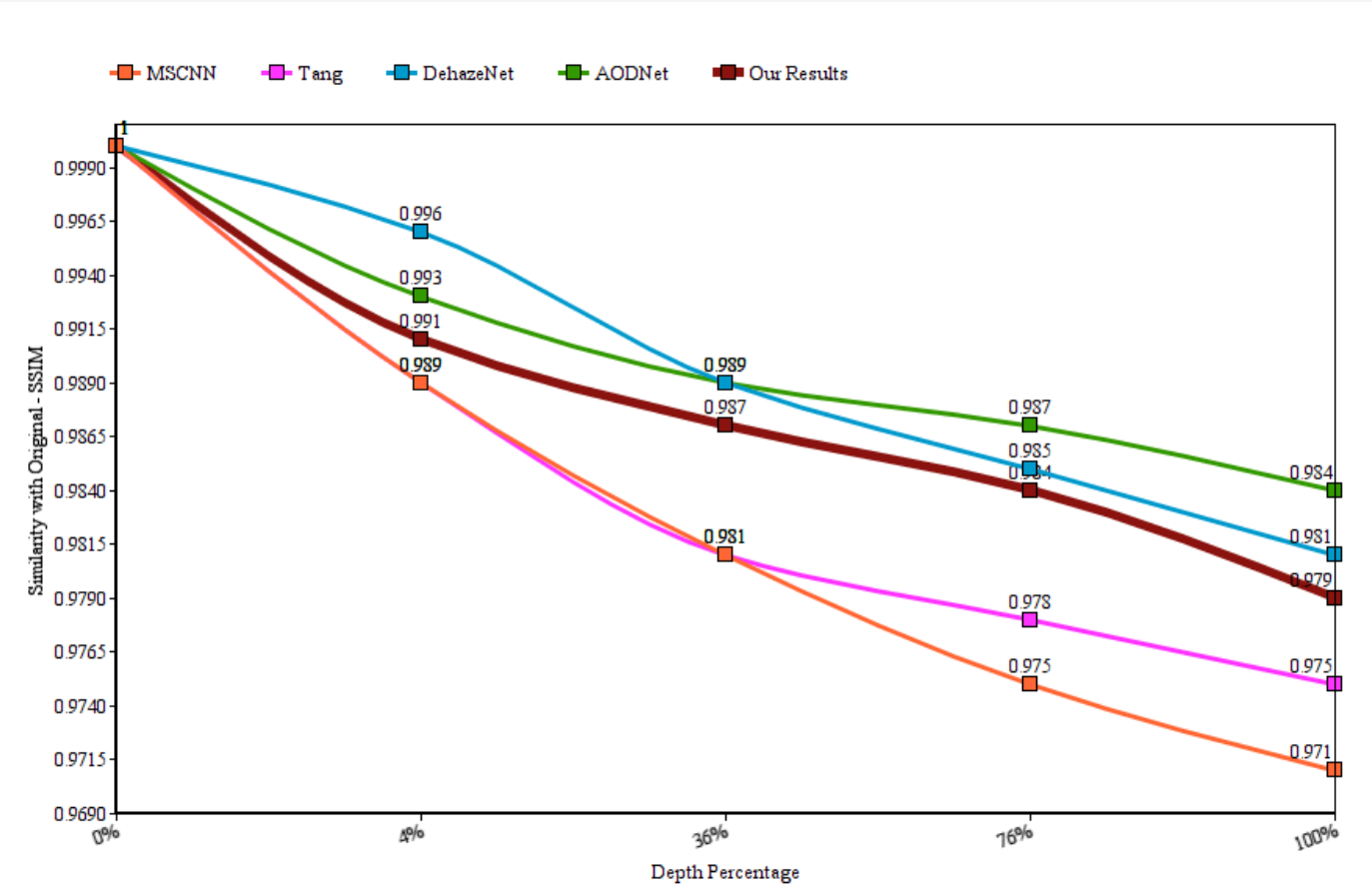}
	\includegraphics[height=2in,width=2.3in]{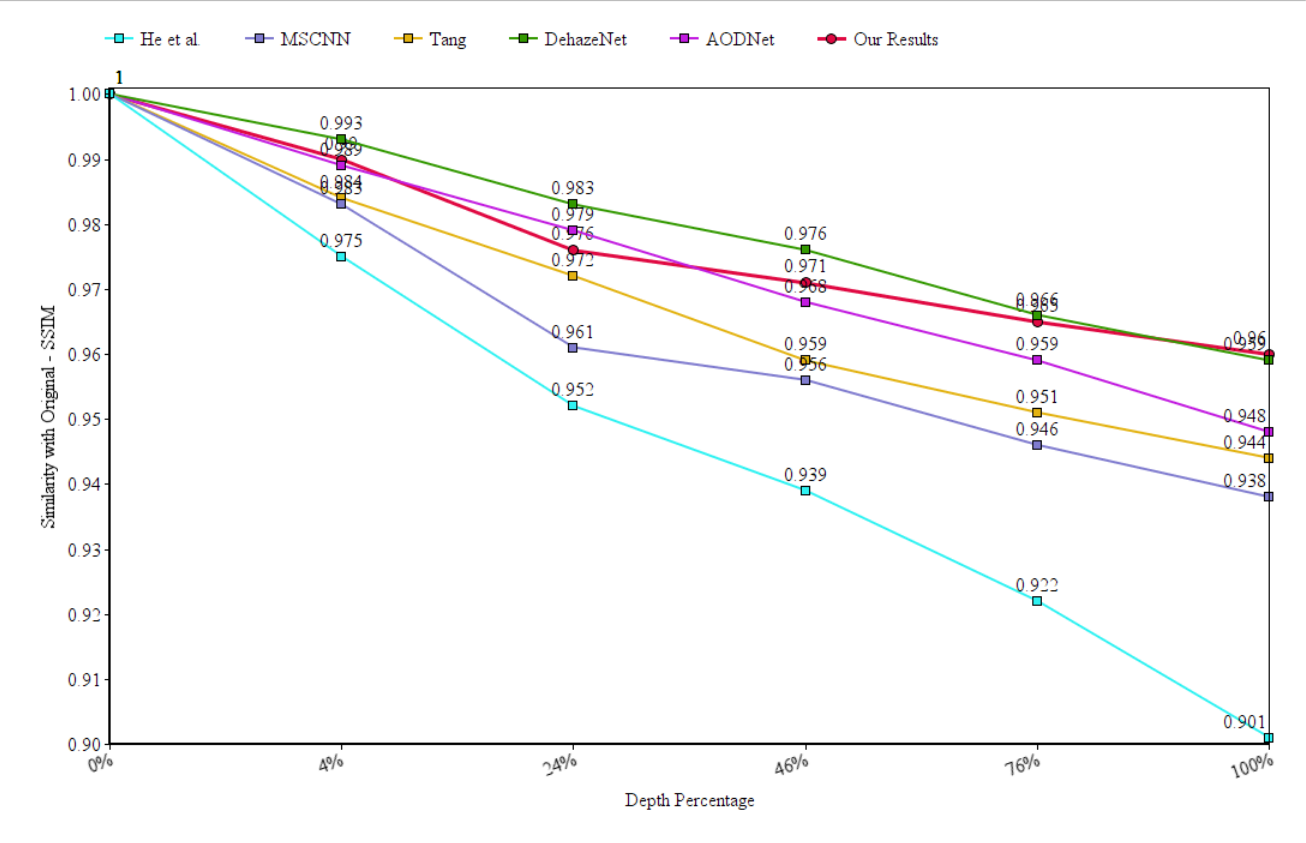}}
\centerline{(a)\hspace{3in}(b)}
\caption{Graphs showing the performance of the proposed approach compared to the state-of-the-art, with increasing depth value. (a) Images without sky area, (b) Images with sky area.}
\end{figure*}

In Table III, the result of implementing the proposed method on images containing sky and other high depth regions, compared to the state-of-the-art methods, are shown. Figure 7 shows how the accuracy of the proposed methods vary with respect to depth. We can observe that, as the depth increases, the proposed method gives much better result compared to the state-of-the-art, including the deep learning based methods.
\begin{figure*}[!htbp]
  \begin{multicols}{6}
  \subfloat[Hazy Image]{
  \includegraphics[scale=0.26]{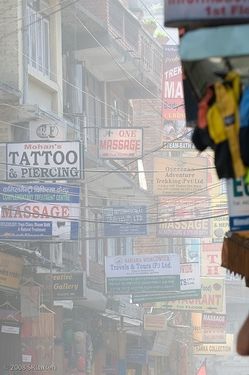}}
  \subfloat[He \textit{et al} {[10]} ]{
  \includegraphics[scale=0.26]{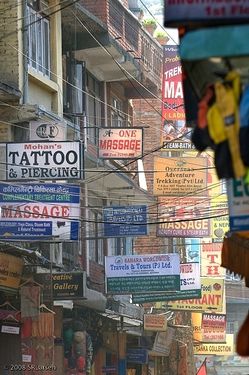}}
  \subfloat[Fattal {\cite{c5}} ]{
  \includegraphics[scale=0.26]{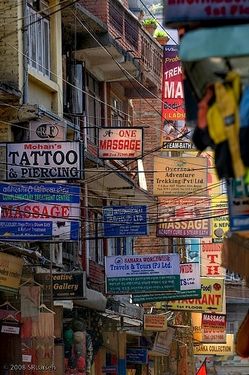}}
  \subfloat[Behat {\cite{c21}}]{
  \includegraphics[scale=0.26]{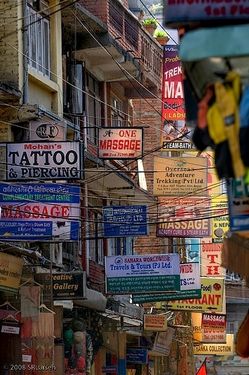}}
  \subfloat[Berman {\cite{c20}}]{
  \includegraphics[scale=0.26]{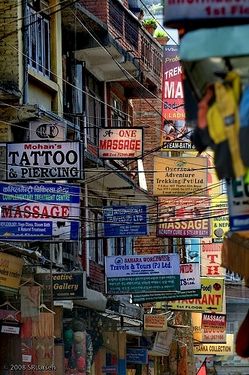}}
  \subfloat[our result]{
  \includegraphics[scale=0.26]{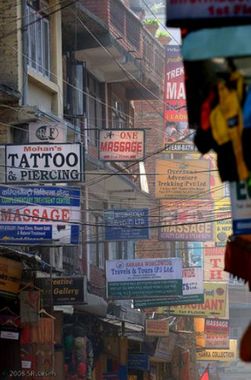}}
  \end{multicols}

  \begin{multicols}{6}
  \subfloat[Hazy Image]{
  \includegraphics[scale=0.31]{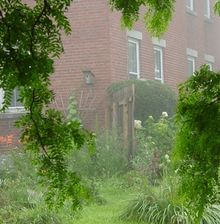}}
  \subfloat[He \textit{et al} {\cite{c10}} ]{
  \includegraphics[scale=0.31]{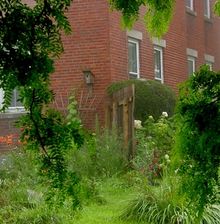}}
  \subfloat[Fattal {\cite{c5}} ]{
  \includegraphics[scale=0.31]{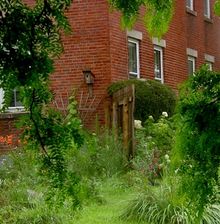}}
  \subfloat[Berman{\cite{c20}}]{
  \includegraphics[scale=0.31]{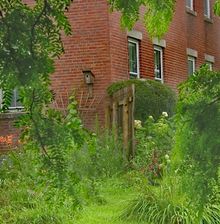}}
  \subfloat[Lu {\cite{c22}} ]{
  \includegraphics[scale=0.51]{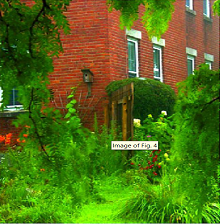}}
  \subfloat[our result]{
  \includegraphics[scale=0.36]{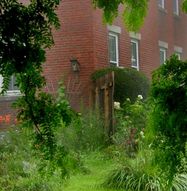}}
  \end{multicols}

  \begin{multicols}{6}
  \subfloat[Hazy Image]{
  \includegraphics[scale=0.11]{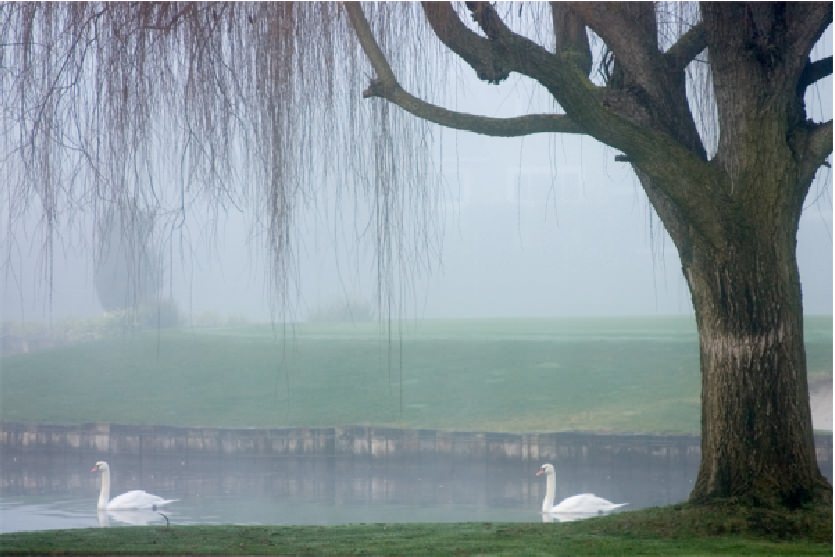}}
  \subfloat[He \textit{et al} {\cite{c10}}]{
  \includegraphics[scale=0.08]{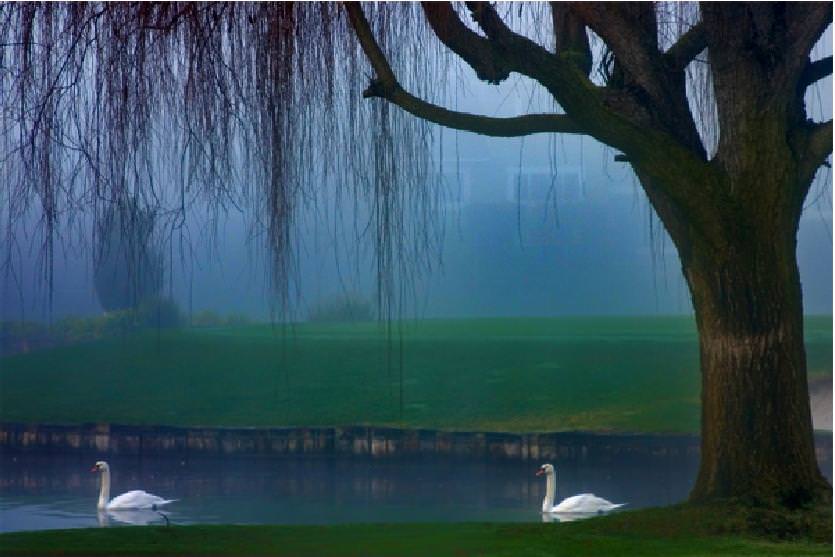}}
  \subfloat[Fattal {\cite{c5}} ]{
  \includegraphics[scale=0.14]{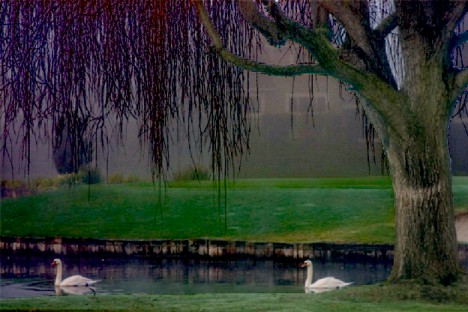}}
  \subfloat[Tang {\cite{c24}}]{
  \includegraphics[scale=0.08]{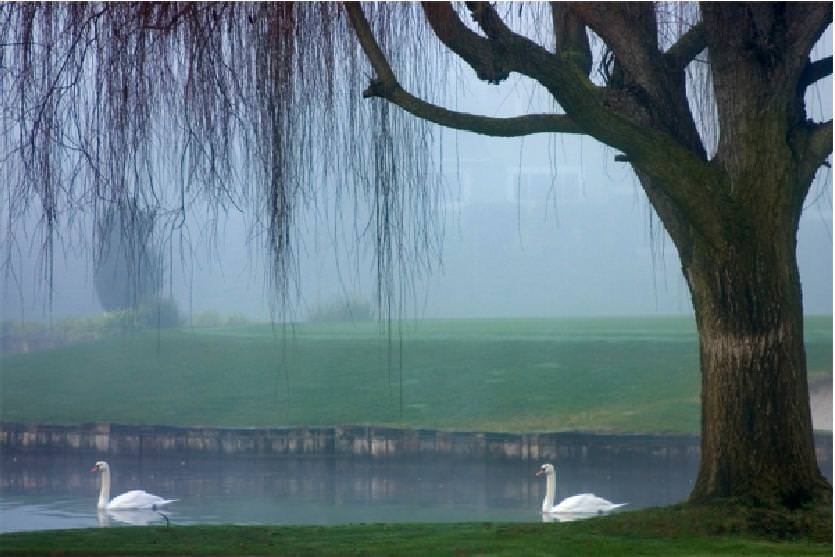}}
  \subfloat[DehazeNet {\cite{c26}}]{
  \includegraphics[scale=0.08]{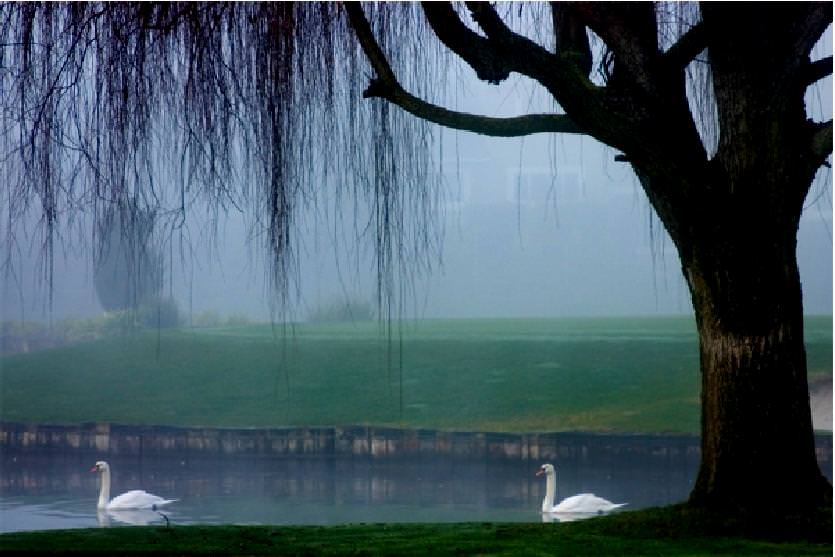}}
  \subfloat[our result]{
  \includegraphics[scale=0.08]{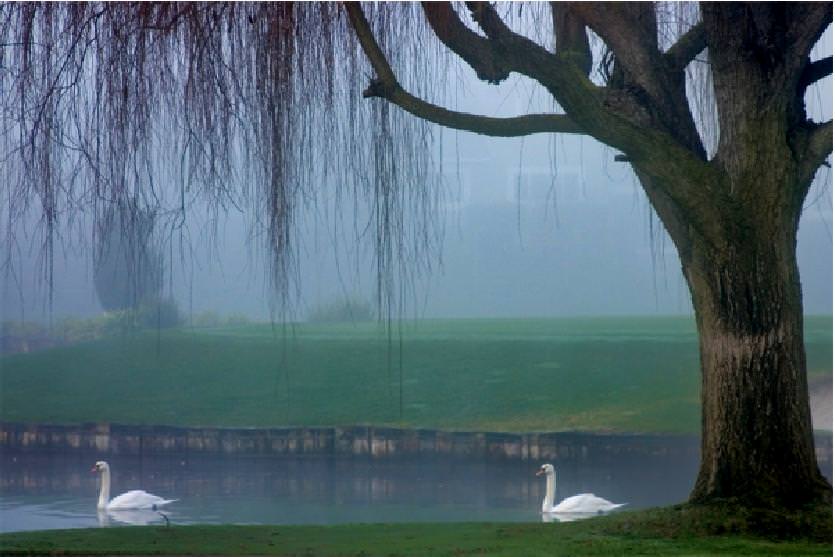}}
  \end{multicols}
\caption{Comparison on natural images: [Left] Input Hazy Images
[Right] Our result. Middle columns display results by several competing methods.}
\end{figure*}

\subsection{Qualitative results}
\label{subsec:qualitative_result}
In Figure 8, we compare the results of the proposed approach with the state-of-the-art single image dehazing techniques \cite{c5,c10,c20,c21,c22}. As noted by \cite{c5}, the picture after fog evacuation may look diminish, since the scene radiance is typically not as splendid as the airlight. The techniques in \cite{c5,c10} provide good results, yet do not have some miniaturized scale differentiate when contrasted with \cite{c20} and to our our method. In the consequence of \cite{c22} there are artefacts in the boundary in between portions.
\begin{figure*}[!htbp]
  \begin{multicols}{6}
  	\subfloat[Hazy Image]{
  	\includegraphics[scale=0.26]{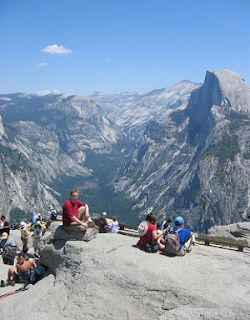}}
 	\subfloat[He \textit{et al} {\cite{c10}}]{
  	\includegraphics[scale=0.26]{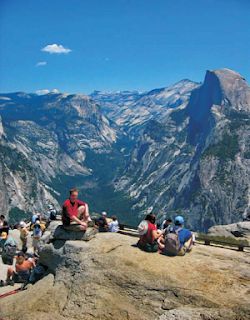}}
  	\subfloat[MSCNN {\cite{c25}}]{
  	\includegraphics[scale=0.11]{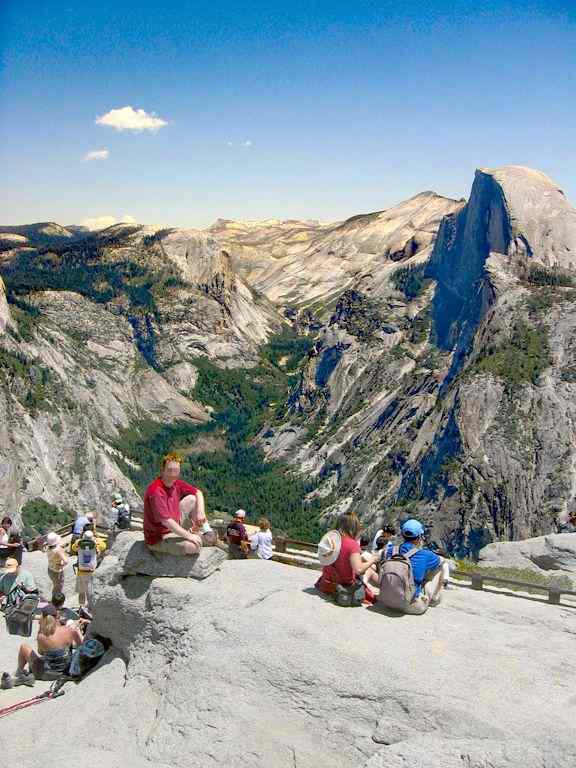}}
  	\subfloat[DehazeNet {\cite{c26}}]{
  	\includegraphics[scale=0.26]{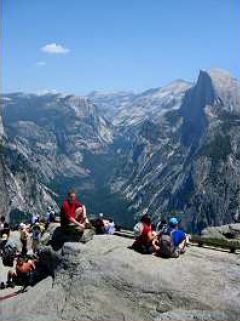}}
  	\subfloat[AODNet {\cite{c27}}]{
  	\includegraphics[scale=0.26]{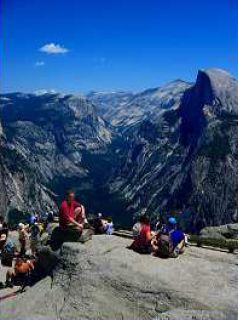}}
  	\subfloat[our result]{
  	\includegraphics[scale=0.26]{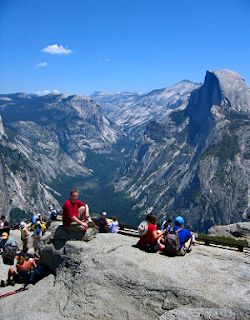}}
  \end{multicols}

  \begin{multicols}{6}
  	\subfloat[Hazy Image]{
  	\includegraphics[scale=0.11]{building_input.jpg}}
  	\subfloat[He \textit{et al} {\cite{c10}}]{
  	\includegraphics[scale=0.11]{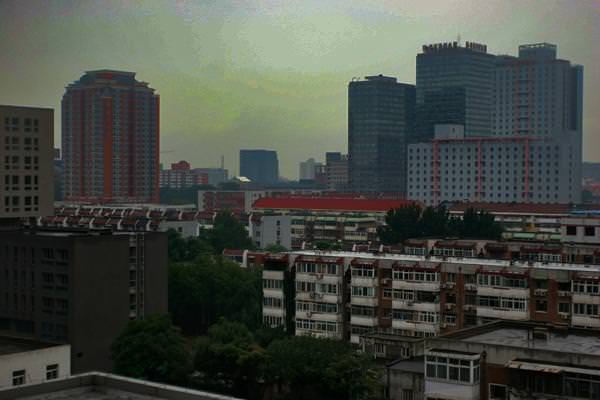}}
  	\subfloat[Tang {\cite{c24}} ]{
  	\includegraphics[scale=0.11]{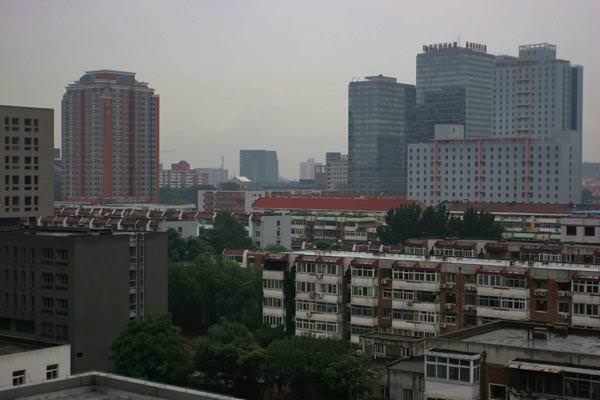}}
  	\subfloat[MSCNN {\cite{c25}}]{
  	\includegraphics[scale=0.33]{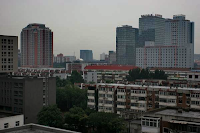}}
  	\subfloat[DehazeNet {\cite{c26}}]{
  	\includegraphics[scale=0.11]{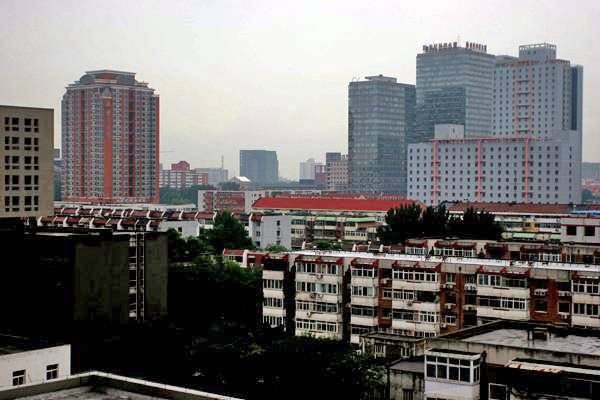}}
  	\subfloat[our result]{
  	\includegraphics[scale=0.11]{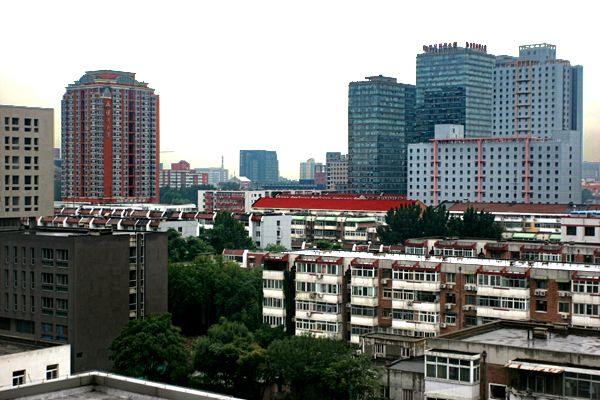}}
  \end{multicols}

	\begin{multicols}{6}
		\subfloat[Hazy Image]{
  		\includegraphics[scale=0.26]{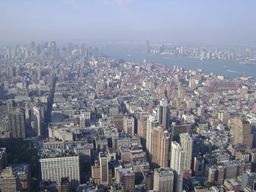}}
  		\subfloat[He \textit{et al} {\cite{c10}}]{
  		\includegraphics[scale=0.26]{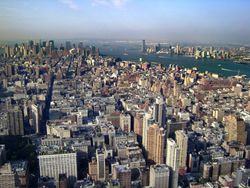}}
  		\subfloat[MSCNN {\cite{c25}}]{
  		\includegraphics[scale=0.26]{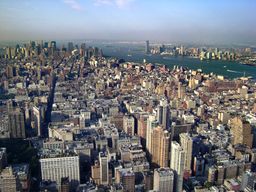}}
  		\subfloat[DehazeNet {\cite{c26}}]{
  		\includegraphics[scale=0.26]{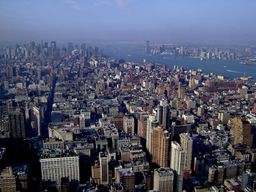}}
  		\subfloat[AODNet {\cite{c27}}]{
  		\includegraphics[scale=0.26]{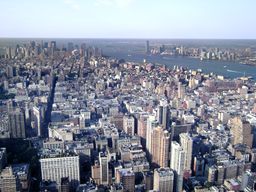}}
  		\subfloat[Our result]{
  		\includegraphics[scale=0.26]{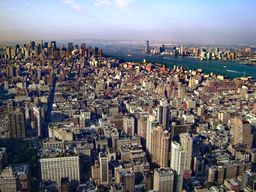}}
	\end{multicols}
	\caption{Comparison on natural images: [Left] Input Hazy Images
[Right] Our result. Middle columns display results by several competing methods.}
\end{figure*}

Figure 9 demonstrates a comparison between the outcomes obtained by the proposed technique and the state-of-the-art  (explicitly, profound deep-learning based techniques) when applied on images containing sky region. The proposed approach outperforms the competing techniques on this kind of images.

Our assumption in regards to having a fog-free pixel in each haze line does not enclose as clear by a few cloudy pixels that set a most extreme radius, e.g. the red structures. In spite of that, the transmission in those territories is evaluated accurately because of the regularization that proliferates the depth data spatially from the other fog lines.

Dehazing the sky area in a hazy image is really challenging, in light of the fact that clouds and mist are similar normal phenomenons with a similar air diffusing illustrate. This issue is facilitated, in any case proceeds in DCP \cite{c10}, Non-Local Dehazing \cite{c20}, DehazeNet \cite{c26} and MSCNN \cite{c25} results. While MSCNN makes the opposite curio of overenhancement: see the sky region of Yosemite for an example (Figure 9). AOD-Net \cite{c27} can oust the obscurity, without displaying fake shading tones or turned dissent shapes. In any case, AOD-Net does not explicitly consider the treatment of white scenes(can be found in sky range of Yosemite and Building in Figure 9). Our method appears to be capable of finding the sky region to keep the shading, and ensures a decent dehazing sway in different locale.

\section{Conclusion}
\label{sec:conclusion}
In this paper, we proposed a method for Image De-Hazing in view of the dim channel earlier and shading lines pixel normality on the regular images. Dark Channel Prior technique depends on insights of the regular outdoor pictures. We got a local nearby formation model that reasons this consistency in foggy/ hazy scenes and portrayed how it is utilized for assessing the scene transmission. We utilize the image forming equation to recuperate haze free picture. We have accepted commitment $A$ to be consistent. We computed transmission and in addition profundity discontinuities. The proposed dehazing scheme delivers comparable outcomes with the recent deep-learning based strategies which require gigantic computational setup. In addition, the proposed technique outperforms the state-of-the-art deep learning based approaches when applied on images with sky regions. Stretching out the recommended model to apply on real-time images may be a conceivable future research heading, which may be extended to apply on hand held devices having less computing power.

\bibliography{mybibfile}

\end{document}